\pdfoutput = 1
\documentclass[twoside,11pt]{article}
\usepackage[preprint]{jmlr2e}

\usepackage[utf8]{inputenc} % allow utf-8 input
\usepackage[T1]{fontenc}    % use 8-bit T1 fonts
\usepackage{url}            % simple URL typesetting
\usepackage{booktabs}       % professional-quality tables
\usepackage{amsfonts}       % blackboard math symbols
\usepackage{nicefrac}       % compact symbols for 1/2, etc.
\usepackage{microtype}      % microtypography
\usepackage{xcolor}         % colors
\usepackage{pifont}         % http://ctan.org/pkg/pifont
\usepackage{breakcites}

\usepackage{array}
\newcommand{\PreserveBackslash}[1]{\let\temp=\\#1\let\\=\temp}
\newcolumntype{C}[1]{>{\PreserveBackslash\centering}p{#1}}
\newcolumntype{R}[1]{>{\PreserveBackslash\raggedleft}p{#1}}
\newcolumntype{L}[1]{>{\PreserveBackslash\raggedright}p{#1}}

\hypersetup{
    colorlinks=true,
    allcolors=brown
}
\usepackage{physics} % shortcuts like norm, abs, qty, etc
\usepackage[linesnumbered,ruled,vlined]{algorithm2e}
\usepackage{xcolor}
\usepackage{amsmath}
\usepackage{amssymb}
\usepackage{numprint}
\usepackage{dsfont}
\usepackage{bm}
\usepackage{bbm}
\usepackage{float}
\usepackage{sectsty}
\usepackage{graphicx} % default *.eps
\usepackage{subcaption}
\usepackage{enumitem}
\usepackage{authblk}
\usepackage{multido}

\newcommand{\R}{\mathbb{R}}
\renewcommand{\P}{\mathbb{P}}
\newcommand{\cmark}{\ding{51}}%
\newcommand{\xmark}{\ding{55}}%

\ShortHeadings{Merlion: A Machine Learning Library for Time Series}{Bhatnagar et al.}
\firstpageno{1}

\title{Merlion: A Machine Learning Library for Time Series}

\author[1]{Aadyot Bhatnagar}
\author[1]{Paul Kassianik}
\author[1]{Chenghao Liu}
\author[1]{Tian Lan}
\author[1]{Wenzhuo Yang}
\author[2]{Rowan Cassius}
\author[1]{Doyen Sahoo}
\author[1]{Devansh Arpit}
\author[2]{Sri Subramanian}
\author[1]{Gerald Woo}
\author[1]{Amrita Saha}
\author[2]{Arun Kumar Jagota}
\author[3]{Gokulakrishnan Gopalakrishnan}
\author[3]{Manpreet Singh}
\author[3]{K C Krithika}
\author[2]{Sukumar Maddineni}
\author[4]{Daeki Cho}
\author[4]{Bo Zong}
\author[1]{Yingbo Zhou}
\author[1]{Caiming Xiong}
\author[1]{Silvio Savarese}
\author[1,*]{Steven Hoi}
\author[1,*]{Huan Wang}
\affil{AI Research, Salesforce}
\affil[2]{Monitoring Cloud, Salesforce}
\affil[3]{Warden AIOps, Salesforce}
\affil[4]{Service Protection, Salesforce}
\affil[*]{Corresponding Authors: \texttt{\{shoi,huan.wang\}@salesforce.com}}

\begin{document}
\maketitle

\begin{abstract}%   <- trailing '%' for backward compatibility of .sty file
We introduce Merlion\protect\footnote{\url{https://github.com/salesforce/Merlion}}, an open-source machine learning library for time series.
It features a unified interface for many commonly used models and datasets for anomaly detection and forecasting on both univariate and multivariate time series, along with standard pre/post-processing layers. It has several modules to improve ease-of-use, including visualization, anomaly score calibration to improve interpetability, AutoML for hyperparameter tuning and model selection, and model ensembling. Merlion also provides a unique evaluation framework that simulates the live deployment and re-training of a model in production.
%It supports anomaly detection and forecasting on both univariate and multivariate time series. It features a unified interface for many commonly used models and datasets, pre/post-processing layers, anomaly score calibration to improve interpretability, AutoML for hyperparameter tuning and model selection, support for ensembles, an evaluation framework that simulates the live deployment \& re-training of a model in production, and a visualization module. 
This library aims to provide engineers and researchers a one-stop solution to rapidly develop models for their specific time series needs and benchmark them across multiple time series datasets.
In this technical report, we highlight Merlion's architecture and major functionalities, and we report benchmark numbers across different baseline models and ensembles.
\end{abstract}


\section{Introduction}

Time series are ubiquitous in monitoring the behavior of complex systems in real-world applications, such as IT operations management, manufacturing industry and cyber security~\citep{hundman2018detecting,Mathur2016,Audibert2020}. They can represent key performance indicators of computing resources such as memory utilization or request latency, business metrics like revenue or daily active users, or feedback for a marketing campaign in the form of social media mentions or ad clickthrough rate. Across all these applications, it is important to accurately forecast the trends and values of key metrics (e.g.\ predicting quarterly sales or planning the capacity required for a server), and to rapidly and accurately detect anomalies in those metrics (e.g.\ an anomalous number of requests to a service can indicate a malicious attack). Indeed, in software industries, anomaly detection, which detects unexpected observations that deviate from normal behaviors and notifies the operators timely to resolve the underlying issues, is one of the critical machine learning techniques to automate the identification of issues and incidents for improving IT system availability in AIOps (AI for IT Operations) \citep{dang2019aiops}. 

Given the wide array of potential applications for time series analytics, numerous tools have been proposed \citep{alibi-detect, kats, seabold2010statsmodels, gluonts_jmlr, law2019stumpy, random_cut_forest, greykite, prophet, pmdarima}. However, there are still many pain points in today’s industry workflows for time series analytics. These include inconsistent interfaces across datasets and models, inconsistent evaluation metrics between academic papers and industrial applications, and a relative lack of support for practical features like post-processing, autoML, and model combination. All of these issues make it challenging to benchmark diverse models across multiple datasets and settings, and subsequently make a data-driven decision about the best model for the task at hand.

This work introduces Merlion, a Python library for time series intelligence. It provides an end-to-end machine learning framework that includes loading and transforming data, building and training models, post-processing model outputs, and evaluating model performance. It supports various time series learning tasks, including forecasting and anomaly detection for both univariate and multivariate time series. Merlion's key features are
\begin{itemize}
    \item Standardized and easily extensible framework for data loading, pre-processing,  and benchmarking for a wide range of time series forecasting and anomaly detection tasks.
    \item A library of diverse models for both anomaly detection and forecasting, unified under a shared interface. Models include classic statistical methods, tree ensembles, and deep learning methods. Advanced users may fully configure each model as desired.
    \item Abstract \texttt{DefaultDetector} and \texttt{DefaultForecaster} models that are efficient, robustly achieve good performance, and provide a starting point for new users (\S\ref{sec:experiments}).
    \item AutoML for automated hyperaparameter tuning and model selection (\S\ref{sec:automl}).
    \item Practical, industry-inspired post-processing rules for anomaly detectors that make anomaly scores more interpretable, while also reducing the false positive rate (\S\ref{sec:post_process}).
    \item Easy-to-use ensembles that combine the outputs of multiple models to achieve more robust performance (\S\ref{sec:anom_ensemble}). 
    \item Flexible evaluation pipelines that simulate the live deployment \& re-training of a model in production (\S\ref{sec:evaluation}), and evaluate performance on both forecasting (\S\ref{sec:forecast_eval}) and anomaly detection (\S\ref{sec:anom_eval}).
    \item Native support for visualizing model predictions.
\end{itemize}

%The rest of the paper is structured as follows: \S\ref{sec:architecture} overviews Merlion's architecture and design principles. \S\ref{sec:forecast} and \S\ref{sec:anomaly} provide algorithmic details of specific Merlion modules for forecasting and anomaly detection, respectively. \S\ref{sec:experiments} includes results from extensive experiments performed using Merlion's evaluation pipeline.

% \TODO{Huan: figure out how to format this properly}
\noindent \textbf{Related Work:}
\pdfoutput = 1
Table \ref{tab:related_works} summarizes how Merlion's key feature set compares with other tools. Broadly, these fall into two categories: libraries which provide unified interfaces for multiple algorithms, such as alibi-detect \citep{alibi-detect}, Kats \citep{kats}, statsmodels \citep{seabold2010statsmodels}, and gluon-ts \citep{gluonts_jmlr}; and single-algorithm solutions, such as Robust Random Cut Forest (RRCF, \cite{random_cut_forest}), STUMPY \citep{law2019stumpy}, Greykite \citep{greykite}, Prophet \citep{prophet}, and pmdarima \citep{pmdarima}.

\begin{table}
    \small
    \centering
    \begin{tabular}{r|c|c|c|c|c|c|c|c}
    \toprule
    & \multicolumn{2}{c|}{Forecast} & \multicolumn{2}{c|}{Anomaly} & AutoML & Ensembles & Benchmarks & Visualization \\
    & Uni & Multi & Uni & Multi & & & \\
    \midrule
    alibi-detect & -- & -- & \cmark & \cmark & -- & -- & -- & -- \\
    Kats & \cmark & \cmark & \cmark & \cmark & \cmark & -- & -- & \cmark  \\
    statsmodels & \cmark & \cmark & -- & -- & -- & -- & -- & --  \\
    gluon-ts & \cmark & \cmark & -- & -- & -- & -- & \cmark & -- \\
    \midrule
    RRCF & -- & -- & \cmark & \cmark & -- & \cmark & -- & -- \\
    STUMPY & -- & -- & \cmark & \cmark & -- & -- & -- & -- \\
    Greykite & \cmark & -- & \cmark & -- & \cmark & -- & -- & \cmark  \\
    Prophet & \cmark & -- & \cmark & -- & -- & -- & -- & \cmark \\
    pmdarima & \cmark & -- & -- & -- & \cmark & -- & -- & -- \\
    \midrule
    Merlion & \cmark & \cmark & \cmark & \cmark & \cmark & \cmark & \cmark & \cmark \\
    \bottomrule
    \end{tabular}
    \caption{Summary of key features supported by Merlion vs.\ other libraries for time series anomaly detection and/or forecasting. First block contains libraries with multiple algorithms; second block contains libraries with a single algorithm. Merlion supports all these features.}
    \label{tab:related_works}
\end{table}

\pdfoutput = 1
\section{Architecture and Design Principles}
\label{sec:architecture}

\begin{figure}[t]
    \centering
    \includegraphics[width=\columnwidth]{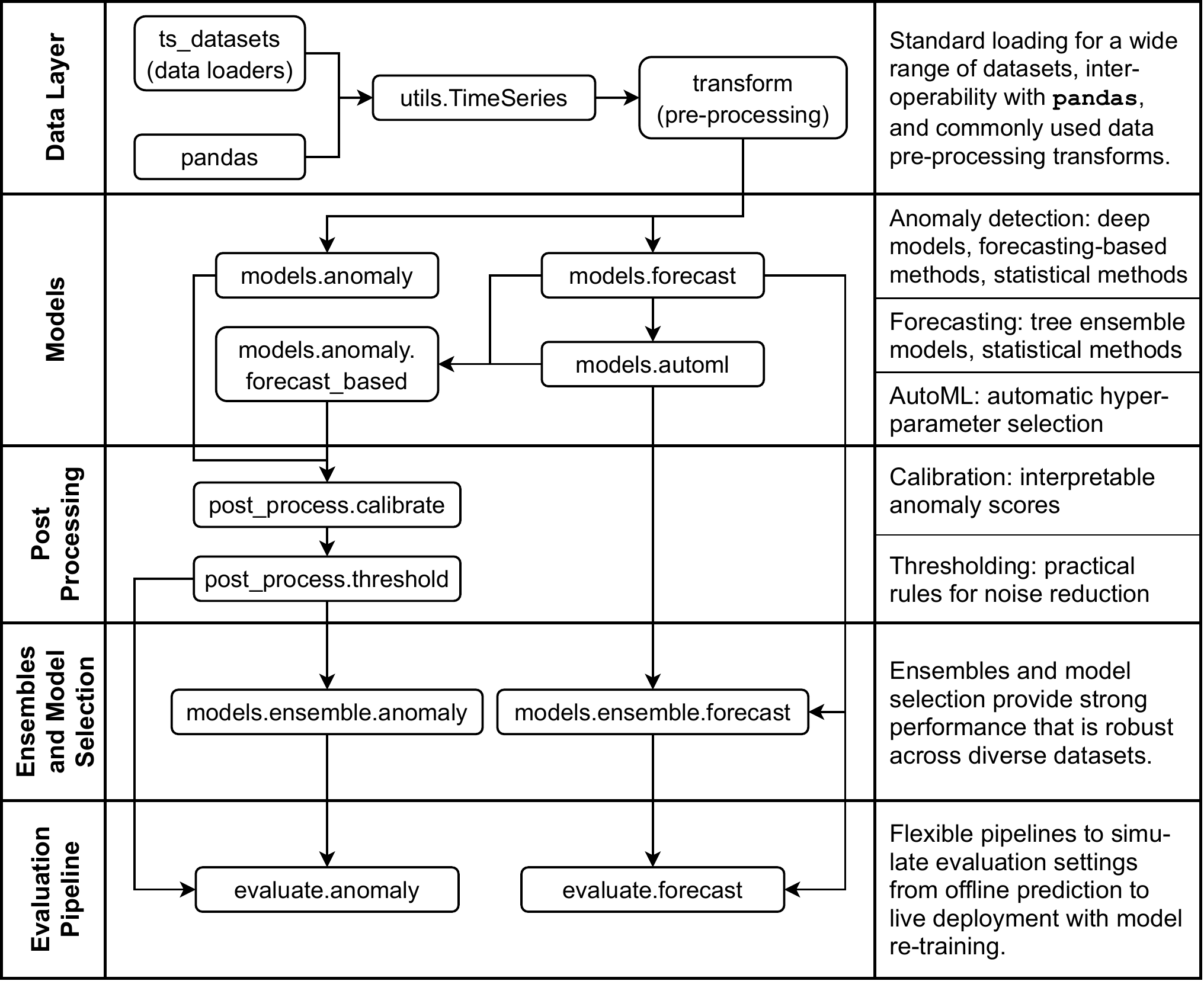}
    \caption{Architecture of modules in Merlion.}
    \label{fig:architecture}
\end{figure}

At a high level, Merlion's module architecture is split into five layers: the data layer loads raw data, converts it to Merlion's \texttt{TimeSeries} data structure, and performs any desired pre-processing; the modeling layer supports a wide array of models for both forecasting and anomaly detection, including autoML for automated hyperparameter tuning; the post-processing layer provides practical solutions to improve the interpetability and reduce the false positive rate of anomaly detection models; the next ensemble layer supports transparent model combination and model selection; and the final evaluation layer implements relevant evaluation metrics and pipelines that simulate the live deployment of a model in production. Figure \ref{fig:architecture} provides a visual overview of the relationships between these modules.

\subsection{Data Layer}
Merlion's core data structure is the \texttt{TimeSeries}, which represents a generic multivariate time series $T$ as a collection of \texttt{UnivariateTimeSeries} $U^{(1)}, \ldots, U^{(d)}$, where each \texttt{UnivariateTimeSeries} is a sequence $U^{(i)} = (t_1^{(i)}, x_1^{(i)}), \ldots, (t_{n_i}^{(i)}, x_{n_i}^{(i)})$. This formulation reflects the reality that individual univariates may be sampled at different rates, or contain missing data at different timestamps. For example, a cloud computing system may report its CPU usage every 10 seconds, but only report the amount of free disk space once per minute.

We allow users to initialize \texttt{TimeSeries} objects directly from \texttt{pandas} dataframes, and we implement standardized loaders for a wide range of datasets in the \texttt{ts\_datasets} package.

Once a \texttt{TimeSeries} has been initialized from raw data, the \texttt{merlion.transform} module supports a host of pre-processing operations that can be applied before passing a \texttt{TimeSeries} to a model. These include resampling, normalization, moving averages, temporal differencing, and others. Notably, multiple transforms can be composed with each other (e.g.\ resampling followed by a moving average), and transforms can be inverted (e.g.\ the normalization $f(x) = (x - \mu) / \sigma$ is inverted as $f^{-1}(y) = \sigma y + \mu$).

\subsection{Models}
Since no single model can perform well across all time series and all use cases, it is important to provide users the flexibility to choose from a  broad suite of heterogenous models. Merlion implements many diverse models for both forecasting and anomaly detection. These include statistical methods, tree-based models, and deep learning approaches, among others. To transparently expose all these options to an end user, we unify all Merlion models under two common API's, one for forecasting, and one for anomaly detection. All models are initialized with a \texttt{config} object that contains implementation-specific hyperparameters, and support a \texttt{model.train(time\_series)} method.

Given a general multivariate time series $T = (U^{(1)}, \ldots, U^{(d)})$, forecasters are trained to predict the values of a single target univariate $U^{(k)}$. One can then obtain a model's forecast of $U^{(k)}$ for a set of future time stamps by calling \texttt{model.forecast(time\_stamps)}.

Analogously, one can obtain an anomaly detector's sequence of anomaly scores for the time series $T$ by simply calling \texttt{model.get\_anomaly\_score(time\_series)}. The handling of univariate vs.\ multivariate time series is implementation-specific (e.g.\ some algorithms look for an anomaly in any of the univariates, while others may look for anomalies in specific univariates). Notably, forecasters can also be used for anomaly detection by treating the residual between the true and predicted value of a target univariate $U^{(k)}$ as an anomaly score. These forecast-based anomaly detectors support both \texttt{model.forecast(time\_stamps)} and \texttt{model.get\_anomaly\_score(time\_series)}.

For models that require additional computation, we implement a \texttt{Layer} interface that is the basis for the autoML features we offer. \texttt{Layer} is used to implement additional logic on top of existing model definitions that would not be properly fitted into the model code itself. Examples include seasonality detection and hyperparameter tuning (see \S\ref{sec:automl} for more details). The \texttt{Layer} is an interface that implements three methods: \texttt{generate\_theta} for generating hyperparameter candidates $ \theta $, \texttt{evaluate\_theta} for evaluating the quality of $\theta$'s, and \texttt{set\_theta} for applying the chosen $\theta$ to the underlying model. A separate class \texttt{ForecasterAutoMLBase} implements \texttt{forecast} and \texttt{train} methods that leverage methods from the \texttt{Layer} class to complete the forecasting model.

Finally, all models support the ability to condition their predictions on historical data \texttt{time\_series\_prev} that is distinct from the data used for training. One can obtain these conditional predictions by calling \texttt{model.forecast(time\_stamps, time\_series\_prev)} or \texttt{model.get\_anomaly\_score(time\_series, time\_series\_prev)}.

\subsection{Post-Processing}
All anomaly detectors have a \texttt{post\_rule} which applies important post-processing to the output of \texttt{model.get\_anomaly\_score(time\_series)}. This includes {\em calibration} (\S\ref{sec:calibration}), which ensures that anomaly scores correspond to standard deviation units and are therefore interpretable and consistent between different models, and {\em thresholding} (\S\ref{sec:thresholding}) rules to reduce the number of false positives. One can directly obtain post-processed anomaly scores by calling \texttt{model.get\_anomaly\_label(time\_series)}.

\subsection{Ensembles and Model Selection}
Ensembles are structured as a model that represents a combination of multiple underlying models. For this purpose we have a base \texttt{EnsembleBase} class that abstracts the process of obtaining predictions $Y_1, \ldots, Y_m$ from $m$ underlying models on a single time series $T$, and a \texttt{Combiner} class that then combines results $Y_1, \ldots, Y_m$ into the output of the ensemble. These combinations include traditional mean ensembles, as well as model selection based on evaluation metrics like sMAPE. Concrete implementations implement the \texttt{forecast} or \texttt{get\_anomaly\_score} methods on top of the tools provided by \texttt{EnsembleBase}, and their \texttt{train} method automatically handles dividing the data into train and validation splits if needed (e.g.\ for model selection).

\subsection{Evaluation Pipeline}
\label{sec:evaluation}
When a time series model is deployed live in production, training and inference are usually not performed in batch on a full time series. Rather, the model is re-trained at a regular cadence, and where possible, inference is performed in a streaming mode. To more realistically simulate this setting, we provide a \texttt{EvaluatorBase} class which implements the following evaluation loop:
\begin{enumerate}
    \item Train an initial model on recent historical training data.
    \item At a regular interval (e.g.\ once per day), retrain the entire model on the most recent data. This can be either the entire history, or a more limited window (e.g.\ 4 weeks).
    \item Obtain the model's predictions (forecasts or anomaly scores) for the time series values that occur between re-trainings. Users may customize whether this should be done in batch, streaming, or at some intermediate cadence.
    \item Compare the model's predictions against the ground truth (actual values for forecasting, or labeled anomalies for anomaly detection), and report quantitative evaluation metrics.
\end{enumerate}
We also provide a wide range of evaluation metrics for both forecasting and anomaly detection, implemented as the enums \texttt{ForecastMetric} and \texttt{TSADMetric}, respectively. Finally, we provide scripts \texttt{benchmark\_forecast.py} and \texttt{benchmark\_anomaly.py} which allow users to use this logic to easily evaluate model performance on any dataset included in the \texttt{ts\_datasets} module. We report experimental results using these scripts in \S\ref{sec:experiments}.

\pdfoutput = 1
\section{Time Series Forecasting}
\label{sec:forecast}
In this section, we introduce Merlion's specific univariate and multivariate forecasting models, provide algorithmic details on Merlion's autoML and ensembling modules for forecasting, and describe the metrics used for experimental evaluation in \S\ref{sec:experiments_uni_forecast} and \S\ref{sec:experiments_multi_forecast}. 

\subsection{Models}
\label{sec:forecast_models}
Merlion contains a number of models for univariate time series forecasting. These include classic statistcal methods like ARIMA (AutoRegressive Integrated Moving Average), SARIMA (Seasonal ARIMA) and ETS (Error, Trend, Seasonality), more recent algorithms like Prophet \citep{prophet}, our previous production algorithm MSES \citep{mses_apidoc}, and a deep autoregressive LSTM \citep{HochSchm97}, among others.

The multivariate forecasting models we use are based on autoregression and tree ensemble algorithms. For the autoregression algorithm, we adopt the Vector Autoregression model (VAR, \cite{multi-var}) that captures the relationship between multiple sequences as they change over time. For tree ensembles, we consider Random Forest (RF, \cite{multi-ho1995random}) and Gradient Boosting (GB, \cite{multi-ke2017lightgbm}) as the base models. However, without appropriate modifications, tree ensemble models can be unsuitable for the general practice of time series forecasting. First, their output is a fixed-length vector, so it is not straightforward to obtain their forecasts for arbitrary prediction horizons. Second, these multivariate forecasting models often have incompatible data workflows or APIs that make it difficult to directly apply them for univariate prediction.

To overcome these obstacles, we propose an autoregressive forecasting strategy for our tree ensemble models, RF Forecaster and GB Forecaster. For a $d$-variable time series, these models forecast the value of all $d$ variables in the time series for time $t_k$, and they condition on this prediction to autoregressively forecast for time $t_{k+1}$. We thus enable these models to produce a forecast for an arbitrary prediction horizon, similar to more traditional models like VAR. Additionally, all our multivariate forecasting models share common APIs with the univariate forecasting models, and are therefore universal for both the univariate and the multivariate forecasting tasks. A full list of supported models can be found in the API documentation.

\subsection{AutoML and Model Selection}
\label{sec:automl}
The AutoML module for time series forecasting models is slightly different from autoML for coventional machine learning models, as we consider not only conventional hyper-parameter optimization, but also the detection of some characteristics of time series. Take the $\mathrm{SARIMA}(p,d,q) \times (P, Q, D)_m$ model as an example. Its hyperparameters include the autoregressive parameter $p$, difference order $d$, moving average parameter $q$, seasonal autoregressive parameter $P$, seasonal difference order $D$, seasonal moving average parameter $Q$, and seasonality $m$. While we use SARIMA as a motivating example, note that automatic seasonality detection can directly enhance other models like ETS and Prophet \citep{prophet}. Meanwhile, choosing the appropriate (seasonal) difference order can yield a representation of the time series that makes the prediction task easier for any model.

Typically, we first analyze the time series to choose the seasonality $m$. Following the idea from the theta method \citep{assimakopoulos2000theta}, we say that a time series has seasonality $m$ at significance level $a$ if
\[
|r_m| > \Phi^{-1}(1-a/2)\sqrt{\frac{1+2\sum^{m-1}_{i=1}r_i^2}{n}},
\]
where $r_k$ is the lag-$k$ autocorrelation, $m$ is the number of periods within a seasonal cycle, $n$ is the sample size, and $\Phi^{-1}$ is the quantile function of the standard normal distribution. By default, we set the significance at $a = 0.05$ to reflect the 95\% prediction intervals.

Next, we select the seasonal difference order $D$ by estimating the strength of the seasonal component with the time series decomposition method \citep{cleveland1990stl}. Suppose the time series can be written as $y = T + S + R$,  where $T$ is the smoothed trend component, $S$ is the seasonal component and $R$ is the remainder. Then, the strength of the seasonality \citep{wang2006characteristic} is 
\[
F_S = \max\left(0, 1- \frac{\mathrm{Var}(R)}{\mathrm{Var}(S+R)}\right).
\]
Note that a time series with strong seasonality will have $F_S$ close to $1$ since $\mathrm{Var}(R)$ is much smaller than $\mathrm{Var}(S+R)$. If $F_S$ is large, we adopt the seasonal difference operation. Thus, we choose $D$ by successviely decomposing the time series and differencing it based on the detected seasonality until $F_S$ is relatively small. We can then choose the difference order $d$ by applying successive KPSS unit-root tests to the seasonally differenced data.

Once $m$, $D$, and $d$ are selected, we can choose the remaining hyperparameters $p$, $q$, $P$, and $Q$ by minimizing the AIC (Akaike Information Criterion) via grid search. Because the parameter space is exponentially large if we exhastively enumerate all the hyperparameter combinations, we follow the step-wise search method of \citet{hyndman2008automatic}. To further speed up the training time of the autoML module, we propose the following approximation strategy: we obtain an initial list of candidate models that achieve good performance after relatively few optimization iterations; we then re-train each of these candidates until model convergence, and finally select the best model by AIC.

\subsection{Ensembles}
\label{sec:forecast_ensemble}
Ensembles of forecasters in Merlion allow a user to transparently combine models in two ways. First, we support traditional ensembles that report the mean or median value predicted by all the models at each timestamp. Second, we support automated model selection. When performing model selection, we divide the training data into train and validation splits, train each model on the training split, and obtain its predictions for the validation split. We then evaluate the quality of those predictions using a user-specified evaluation metric like sMAPE or RMSE (\S\ref{sec:forecast_eval}), and return the model that achieved the best performance after re-training it on the full training data. These features are useful in many practical scenarios, as they can greatly reduce the amount of human intervention needed when deploying a model.

\subsection{Evaluation Metrics}
\label{sec:forecast_eval}
There are many ways to evaluate the accuracy of a forecasting model. Given a time series with $n$ observations, let $y_t$ denote the observed value at time $t$ and $\hat{y}_t$ denote the corresponding predicted value. Then the forecasting error $e_t$ at time $t$ is $y_t-\hat{y}_t$. Error measures based on absolute or squared errors are widely used, and they are formuated as \[\text{MAE}=\frac{1}{n}\sum^n_{t=1}|e_t|\quad  \text{and} \quad \text{RMSE} = \sqrt{\frac{\sum^n_{t=1}e_t^2}{n}},\]
respectively. Unfortunately, these measures cannot be compared across time series that are on different scales. To achive scale-independence, one alternative approach is to use percentage errors based on the observed values. One typical measure is sMAPE \citep{makridakis2000m3}, defined as:
\[\text{sMAPE} = \frac{1}{n}\sum^n_{t=1}\frac{|e_t|}{|y_t|+|\hat{y}_t|}*200(\%),\] Unfortunately, sMAPE has the disadvantage of being ill-defined if $y_t$ and $\hat{y}_t$ are both close to zero. To address this issue, one alternative measure is MARRE, which  is defined as
\[\text{MARRE}=\frac{1}{n}\frac{\sum^n_{t=1}|e_t|}{|\max(y_t) - \min(y_t)|}.\]
However, MARRE is not always suitable for non-stationary time series, where the scale of the data may evolve over time.
%Another commonly used one is MASE \citep{hyndman2006another}, which scales the errors based on the in-sample MAE from the naive method. Here, naive method means to generate one-period-ahead forecasts from each data point in the sample. Formally, it is defined as 
%\[\text{MASE}=\frac{1}{n}\frac{\sum^{n_0+n}_{t=n_0+1}|e_t|}{\frac{1}{n_0-m}\sum^{n_0}_{t=m+1}|y_{t} - y_{t-m}|}.\]
%Here, $n_0$ denotes the number of samples in the training set, $n$ denotes the number of samples in the testset and $m$ denote the period. Finally, understanding the uncertainty of a model's forecast can be useful for decision makers in various domains. An alternative measure to evaluate forecastiing uncertainty inerval is to use MSIS \citep{gneiting2007strictly}, which is defined as
%\[\text{MSIS} = \frac{1}{n}\frac{\sum^{n_0+n}_{t=n_0+1}(U_t-L_t)+\frac{2}{a}(L_t-y_t)\mathbb{I}[y_t < L_t] + \frac{2}{a}(y_t - U_t)\mathbb{I}[y_t > U_t]}{\frac{1}{n_0-m}\sum^{n_0}_{t=m+1}|y_t - y_{t-m}|}.\]
%Here, $L_t$ and $U_t$ are the lower and upper bounds of the prediction intervals, $a$ is the significance level ($a$ is set to $0.05$ by default to reflect the $95\%$ prediction intervals), and $\mathbb{I}[\cdot]$ is the indicator function.
Merlion's \texttt{ForecastMetric} enum supports all the above classes of evaluation metrics, as well as others detailed in the API documentation. By default, we evaluate forecasting results by sMAPE, but users may manually specify alternatives if their application calls for it.

\pdfoutput = 1
\section{Time Series Anomaly Detection}
\label{sec:anomaly}
In this section, we introduce Merlion's specific univariate and multivariate anomaly detection models, provide algorithmic details on Merlion's post-processing and ensembling modules for anomaly detection, and describe the metrics used for experimental evaluation in \S\ref{sec:experiments_uni_anom} and \S\ref{sec:experiments_multi_anom}. 

\subsection{Models}
\label{sec:anom_models}
Merlion contains a number of models that are specialized for univariate time series anomaly detection. These fall into two groups: forecasting-based and statistical. Because forecasters in Merlion predict the value of a specific univariate in a general time series, they are straightforward to adapt for anomaly detection. The anomaly score is simply the residual between the predicted and true time series value, optionally normalized by the underlying forecaster's predicted standard error (if it produces one).

For univariate statistical methods, we support Spectral Residual \citep{spectral_residual}, as well as two simple baselines WindStats and ZMS. WindStats divides each week into windows (e.g.\ 6 hours), and computes the anomaly score $s_t = (x_t - \mu(t)) / \sigma(t)$, where $\mu(t)$ and $\sigma(t)$ are the historical mean and standard deviation of the window in question (e.g.\ 12pm to 6pm on Monday). ZMS computes $k$-lags $\Delta^{(k)}_t = x_{t} - x_{t-k}$ for $k = 1, 2, 4, 8, \ldots$, and computes the anomaly score $s_t = \max_k (\Delta^{(k)}_t - \mu^{(k)}) / \sigma^{(k)}$, where $\mu^{(k)}$ and $\sigma^{(k)}$ are the mean and standard deviation of each $k$-lag.

In addition to models that are specialized for univariate anomaly detection, we support both statistical methods and deep learning models that can handle both univariate and multivariate anomaly detection. The statistical methods include Isolation Forest \citep{isolation_forest} and Random Cut Forest \cite{random_cut_forest}, while the deep learning models include an autoencoder \citep{autoencoder}, a Deep Autoencoding Gaussian Mixture Model (DAGMM, \cite{dagmm}), a LSTM encoder-decoder \citep{lstmed}, and a variational autoencoder \citep{vae}. A full list can be found in the API documentation.

\subsection{Post-Processing}
\label{sec:post_process}
Merlion supports two key post-processing steps for anomaly detectors: calibration and thresholding. Calibration is important for improving model intepretability, while thresholding converts a sequence of continuous anomaly scores into discrete labels and reduces the false positive rate. 

\subsubsection{Calibration}
\label{sec:calibration}
All anomaly detectors in Merlion return anomaly scores $s_t$ such that $\abs{s_t}$ is positively correlated with the severity of the anomaly. However, the scales and distributions of these anomaly scores vary widely. For example, Isolation Forest \citep{isolation_forest} returns an anomaly score $s_t \in [0, 1]$ where $-\log_2(1 - s_t)$ is a node's depth in a binary tree; Spectral Residual \citep{spectral_residual} returns an unnormalized saliency map; DAGMM \citep{dagmm} returns a negative log probability.

To successfully use a model, one must be able to interpret the anomaly scores it returns. However, this prevents many models from being immediately useful to users who are unfamiliar with their specifc implementations. Calibration bridges this gap by making all anomaly scores interpretable as z-scores, i.e.\ values drawn from a standard normal distribution.

Let $\Phi: \R \to [0, 1]$ be the cumulative distribution function (CDF) of the standard normal distribution. A calibrator $C: \R \to \R$ converts a raw anomaly score $s_t$ into a calibrated score $z_t$ such that $\P[\abs{z_t} > \alpha] = \Phi(\alpha) - \Phi(-\alpha) = 2 \Phi(\alpha) - 1$, i.e.\ $\abs{z_t}$ follows the same distribution as the absolute value of a standard normal random variable. If $F_s: \mathbb{R} \to [0, 1]$ is the CDF of the absolute raw anomaly scores $\abs{s_t}$, we can compute the recalibrated score as
\[C(s_t) = \mathrm{sign}(s_t) \Phi^{-1}((1 + F_s(\abs{s_t})) / 2) \]
Notably, this is also the optimal transport map between the two distributions, i.e.\ the mapping which transfers mass from one distribution to the other in a way that minimizes the $\ell_1$ transportation cost \citep{villani2003topics}.

In practice, we estimate the calibration function $C$ using the empirical CDF $\hat{F}_s$ and a monotone spline interpolator \citep{pchip} for intermediate values. This simple post-processing step dramatically increases the intepretability of the anomaly scores returned by individual models. It also enables us to create ensembles of diverse anomaly detectors as discussed in \S\ref{sec:anom_ensemble}. This functionality is enabled by default for all anomaly detection models.

\subsubsection{Thresholding}
\label{sec:thresholding}
The most common way to decide whether an individual timestamp $t$ as anomalous, is to compare the anomaly score $s_t$ against a threshold $\tau$. However, in many real-world systems, a human is alerted every time an anomaly is detected. A high false positive rate will lead to fatigue from the end users who have to investigate each alert, and may result in a system whose users think it is unreliable. 

A common way to circumvent this problem is including additional automated checks that must pass before a human is alerted. For instance, our system may only fire an alert if there are {\em two} timestamps $t_1, t_2$ within a short window (e.g.\ 1 hour) with a high anomaly score, i.e.\ $\abs{s_{t_1}} > \tau$ {\em and} $\abs{s_{t_2}} > \tau$. Moreover, if we know that anomalies typically last 2 hours, our system can simply suppress all alerts occurring within 2 hours of a recent one, as those alerts likely correspond to the same underlying incident. Often, one or both of these steps can greatly increase precision without adversely impacting recall. As such, they are commonplace in production systems.

Merlion implements all these features in the user-configurable \texttt{AggregateAlarms} post-processing rule, and by default, it is enabled for all anomaly detection models.

\subsection{Ensembles}
\label{sec:anom_ensemble}
Because both time series and the anomalies they contain are incredibly diverse, it is unlikely that a single model will be the best for all use cases. In principle, a heterogenous ensemble of models may generalize better than any individual model in that ensemble. Unfortunately, as mentioned in \S\ref{sec:calibration}, constructing an ensemble is not straightforward due to the vast differences in anomaly scores returned by various models.

However, because we can rely on the anomaly scores of all Merlion models to be interpretable as z-scores, we can construct an ensemble of anomaly detectors by simply reporting the mean {\em calibrated} anomaly score returned by each individual model, and then applying a threshold (a la \S\ref{sec:thresholding}) on this combined anomaly score. Empirically, we find that ensembles robustly achieve the strongest or competitive performance (relative to the baselines considered) across multiple open source and internal datasets for both univariate (Table \ref{tab:uni_anom_full_results}) and multivariate (Table \ref{tab:multi_anom_full_results}) anomaly detection.

\subsection{Evaluation Metrics}
\label{sec:anom_eval}
The key challenge in designing an appropriate evaluation metric for time series anomaly detection lies in the fact that anomalies are almost always windows of time, rather than discrete points. Thus, while it is easy to compute the pointwise (PW) precision, recall, and F1 score of a predicted anomaly label sequence relative to the ground truth label sequence, these metrics do not reflect a quantity that human operators care about.

\cite{xu2018unsupervised} propose {\em point-adjusted} (PA) metrics as a solution to this problem: if any point in a ground truth anomaly window is labeled as anomalous, all points in the segment are treated as true positives. If no anomalies are flagged in the window, all points are labeled as false negatives. Any predicted anomalies outside of anomalous windows are treated as false positives. Precision, recall, and F1 can be computed based on these adjusted true/false positive/negative counts. However, the disadvantage of PA metrics is that they are biased to reward models for detecting long anomalies more than short ones. 

\cite{hundman2018detecting} propose {\em revised point-adjusted} (RPA) metrics as a closely related alternative: if any point in a ground truth anomaly window is labeled as anomalous, {\em one} true positive is registered. If no anomalies are flagged in the window, {\em one} false negative is recorded. Any predicted anomalies outside of anomalous windows are treated as false positives. These metrics address the shortcomings of PA metrics, but they do penalize false positives more heavily than alternatives.

Merlion's \texttt{TSADMetric} enum supports all 3 classes of evaluation metrics (PW, PA, and RPA), as well as the mean time to detect an anomaly. By default, we evaluate RPA metrics, but users may manually specify alternatives if their application calls for it.

\pdfoutput = 1
\section{Experiments}
\label{sec:experiments}
In the experiments section, we show benchmark results generated by using Merlion with popular baseline models across several time series datasets. The purpose of this section is not to obtain state-of-the-art results. Rather, most algorithms listed are strong baselines. To avoid the possible risk of label leaking through manual hyperparameter tuning, for all experiments, we evaluate all models with a single choice of sensible default hyperparameters and data pre-processing, regardless of dataset. 

\subsection{Univariate Forecasting}
\label{sec:experiments_uni_forecast}

\subsubsection{Datasets and Evaluation}
We primarily evaluate our models on the M4 benchmark \citep{makridakis2018m4, m42018m4}, an influential time series forecasting competition. The dataset contains $100,000$ time series from diverse domains including financial, industry, and demographic forecasting. It has sampling frequencies ranging from hourly to yearly. Table \ref{tab:m4} summarizes the dataset. We additionally evaluate on three internal datasets of cloud KPIs, which we describe in Table \ref{tab:uni_forecast_internal_data}. For all datasets, models are trained and evaluated in the offline batch prediction setting, with a pre-defined prediction horizon equal to the size of the test split. To mitigate the effect of outliers, we report both the mean and median sMAPE for each method.

\begin{table}
    \small
    \centering
    \begin{tabular}{r|c|c|c|c|c|c}
        \toprule
        M4 & Hourly & Daily & Weekly & Monthly & Quarterly & Yearly\\ \midrule
        \# Time Series & 414 & 4,227 & 359 & 48,000 & 24,000 & 23,000 \\
        %Mean Length & 902 & 2371 & 1035 & 234 & 100 & 37 \\
        %SD Length & 128 & 1757 & 707 & 137 & 51 & 24 \\
        Real Data? & \cmark & \cmark & \cmark & \cmark & \cmark & \cmark \\
        Availability & Public & Public & Public & Public& Public& Public\\
        \midrule
        Prediction Horizon & 6 & 8 & 18 & 13 & 14 & 48\\
        \bottomrule
    \end{tabular}
    \caption{Summary of M4 dataset for univariate forecasting.}
    \label{tab:m4}
\end{table}

\begin{table}
    \small
    \centering
    \begin{tabular}{r|c|c|c}
        \toprule
        & Int\_UF1 & Int\_UF2 & Int\_UF3 \\ 
        \midrule
        \# Time Series & 21 &  6 & 4 \\
        %Mean Length & 9374 & 557 & 4420 \\
        %SD Length & 0.71& 171 & 2914\\
        Real Data? & \cmark & \cmark & \cmark \\
        Availability & Internal & Internal & Internal \\
        \midrule
       Train Split & First 75\% & First 25\% & First 25\%\\
        \bottomrule
    \end{tabular}
    \caption{Summary of internal datasets for univariate forecasting.}
    \label{tab:uni_forecast_internal_data}
\end{table}

\subsubsection{Models}
We compare ARIMA (AutoRegressive Integrated Moving Average), Prophet \citep{prophet}, ETS (Error, Trend, Seasonality), and MSES (the previous production solution, \cite{mses_apidoc}). For ARIMA, Prophet and ETS, we also consider our autoML variants that perform automatic seasonality detection and hyperparameter tuning, as described in \S\ref{sec:automl}. These are implemented using the \texttt{merlion.models.automl} module.
\begin{table}
    \small
    \centering
    \begin{tabular}{r|c|c|c|c|c|c}
        \toprule
        & \multicolumn{2}{c}{M4$\_$Hourly} & \multicolumn{2}{|c|}{M4$\_$Daily} & \multicolumn{2}{c}{M4$\_$Weekly} \\\hline
        sMAPE& Mean & Median & Mean & Median & Mean & Median \\ \hline
        MSES & 32.45 & 16.89 & 5.87 & 3.97 & 16.53 & 9.96 \\
        ARIMA & 33.54 & 19.27 & 3.23 & 2.06 & 9.29 & 5.38 \\
        Prophet & 18.08 & 6.92 & 11.67 & 5.99  & 19.98 & 11.26 \\
        ETS & 42.95 & 19.88 & \textbf{3.04} & 2.00 & 9.00 & 5.17 \\
        \midrule
        AutoSARIMA & \textbf{13.61} & \textbf{4.73} & 3.29 & 2.00 & \textbf{8.30} & \textbf{5.09} \\
        AutoProphet & 16.49 & 6.20 & 11.67 & 5.98 & 20.01 & 11.63 \\
        AutoETS & 19.23 & 5.33 & 3.07 & \textbf{1.98} & 9.32 & 5.15 \\
        \bottomrule
    \end{tabular}
    
    \begin{tabular}{r|c|c|c|c|c|c}
        \toprule
        & \multicolumn{2}{c}{M4$\_$Monthly} & \multicolumn{2}{|c|}{M4$\_$Quarterly} & \multicolumn{2}{c}{M4$\_$Yearly} \\\hline
        sMAPE& Mean & Median & Mean & Median & Mean & Median \\ \hline
        MSES & 25.40 & 13.69 & 19.03 & 9.53 & 21.63 & 12.11 \\
        ARIMA & 17.66 & 10.41 & 13.37 & 8.27 & 16.37 & 10.34 \\
        Prophet & 20.64 & 11.00 & 24.53 & 12.81  & 30.23 & 19.24 \\
        ETS & 14.32 & 8.37 & 11.08 & 6.94 & 16.43 & 11.43 \\
        \midrule
        AutoSARIMA & 14.26 & \textbf{7.17} & 10.51 & \textbf{5.44} & 17.16 & 9.48 \\
        AutoProphet & 20.43 & 10.42 & 24.62 & 12.87 & 30.23 & 19.24 \\
        AutoETS & \textbf{13.73} & 7.41 & \textbf{10.33} & 5.77 & \textbf{15.96} & \textbf{9.18} \\
        \bottomrule
    \end{tabular}
    \caption{Mean and Median sMAPE achived by univariate forecasting models on M4 datasets. All models were evaluated without retraining. Best results are in \textbf{bold}.}
    \label{tab:uni_forecast_result_public}
\end{table}

\begin{table}
    \small
    \centering
    \begin{tabular}{r|c|c|c|c|c|c}
        \toprule
        & \multicolumn{2}{c}{Int\_UF1} & \multicolumn{2}{|c|}{Int\_UF2} & \multicolumn{2}{c}{Int\_UF3} \\\hline
        sMAPE& Mean & Median & Mean & Median & Mean & Median \\ \hline
        MSES & 33.50 & 32.60 & 32.30 & 14.45 & 3.882 & 3.771 \\
        ARIMA & 24.00 & 19.42 & 25.89 & 24.40 &  5.94 & 6.15\\
        Prophet & 56.07 & 42.10 & 30.93 & 38.54 & 72.55 & 78.27 \\
        ETS & 17.02 & 16.36 & 25.10 & 24.19 & 3.55 & 3.32 \\
        \midrule
        AutoSARIMA & \textbf{15.98} & \textbf{15.10} & \textbf{17.45} & \textbf{9.98} & 3.42 & 3.41 \\
        AutoProphet & 56.43 & 41.99 & 26.35 & 24.19 & 69.32 & 78.37 \\
        AutoETS & 21.65 & 20.15 & 19.41 & 12.67 & \textbf{3.15} & \textbf{2.94} \\
        \bottomrule
    \end{tabular}
    \caption{Mean and Median sMAPE achived by univariate forecasting models on internal datasets. All models were evaluated without retraining. Best results are in \textbf{bold}.}
    \label{tab:uni_forecast_result_internal}
\end{table}

\begin{table}
    \small
    \centering
    \begin{tabular}{r|c}
        \toprule
         ARIMA $\to$ AutoSARIMA & $5.03$ ($p=0.010$) \\
         Prophet $\to$ AutoProphet & $1.01$ ($p=0.035$) \\
         ETS $\to$ AutoETS & $2.96$ ($p=0.039$) \\
         \bottomrule
    \end{tabular}
    \caption{Average reduction (improvment) in sMAPE achieved by applying autoML. $p$-value is from a 2-sided paired sample $t$-test. Our autoML module improves all 3 models at significance level $p = 0.05$.
    }
    \label{tab:automl_results}
\end{table}

\subsubsection{Results}

Table \ref{tab:uni_forecast_result_public} and \ref{tab:uni_forecast_result_internal} respectively show the performance of each model on the public and internal datasets. Table \ref{tab:automl_results} shows the average improvement achieved by using the autoML module. For ARIMA, we find a clear improvement by using the AutoML module for all datasets. While the improvement is statistically significant for Prophet and AutoETS overall, the actual change is small on most datasets except M4\_Hourly and Int\_UF2. This is because many of the other datasets don't contain well-defined seasonalities; additionally, Prophet already has some pre-defined seasonality detection (daily, weekly, and yearly), but these datasets contain time series with hourly seasonalities.

While there is no clear winner across all datasets, AutoSarima and AutoETS consistently outperform other methods. The overall performance of AutoSarima is slightly better than AutoETS, but AutoSarima is much slower to train. Therefore, we believe that AutoETS is a good ``default'' model for new users or early exploration.

\subsection{Multivariate Forecasting}
\label{sec:experiments_multi_forecast}

\subsubsection{Datasets and Evaluation}
\begin{table}
    \small
    \centering
    \begin{tabular}{r|c|c|c|c}
        \toprule
        & Power Grid & Seattle Trail & Solar Plant & Int\_MF \\ \midrule
        \# Time Series & 1 & 1 & 1 & 21\\
        \# Variables & 10 & 5 & 405 & 22 \\
        Real Data? & \cmark & \cmark & \cmark & \cmark\\
        Availability & Public & Public & Public & Internal\\
        Data Source & Energy Demand & Trail Traffic & Power Generation & Cloud KPIs \\ \midrule
        Train Split & First 70\% & First 70\% & First 70\%  & First 75\% \\
        Granularity & 1h & \xmark & 30min & 10s\\
        \midrule
        Reference & \cite{energy_consumption_dataset} & \cite{seattle_dataset} & \cite{solar_power_dataset} & -- \\
        \bottomrule
    \end{tabular}
    \caption{Summary of multivariate forecasting datasets.}
    \label{tab:multi_forecast_data}
\end{table}
We collect public and internal datasets (Table \ref{tab:multi_forecast_data}) and train models on a training split of the data. For some datasets, we resample the data with given granularities. For each time series, we train the models on the training split to predict the {\em first} univariate as a target sequence. While we don't re-train the model, we use the evaluation pipeline described in \S\ref{sec:evaluation} to incrementally obtain predictions for the test split using a rolling window. We predict time series values for the next 3 timestamps, while conditioning the prediction on the previous 21 timestamps. We obtain these 3-step predictions at every timestamp in the test split, and evaluate the quality of the prediction using sMAPE where possible, and RMSE otherwise.

\subsubsection{Models}
The multivariate forecasting models we use are based on the autoregression and tree ensemble algorithms. We compare VAR \citep{multi-var}, GB Forecaster based on the Gradient Boosting algorithm \citep{multi-ke2017lightgbm} and RF Forecaster based on the Random Forest algorithm \citep{multi-ho1995random}. We discuss the implementation details of these models in \S\ref{sec:forecast_models}. For the GB Forecster and RF Forecaster, we use our proposed autoregression strategy to enable them to forecast for an arbitrary prediction horizon.

\subsubsection{Results}
Table \ref{tab:multi_forecast_result} reports the performance of each model. We find that our proposed autogression-based tree ensemble model GB Forecaster achieves the best results on three of the four datasets and is competitive with the best result on the fourth. While the VAR model shows competitive performance on some datasets, it is not as robust. For example, the Seattle Trail dataset has large outliers (several orders of magnitude larger than the mean) that dramatically impact the performance of VAR. For this reason, we believe that GB Forecaster is a good ``default'' model for new users or early exploration.

\begin{table}
    \small
    \centering
    \begin{tabular}{r|c|c|c|c|c}
        \toprule
        & Power Grid & Seattle Trail & Solar Plant & \multicolumn{2}{c}{Int\_MF} \\ \midrule
        & sMAPE & RMSE & RMSE & sMAPE & sMAPE \\
        & (1 TS) & (1 TS) & (1 TS) & (mean) & (median) \\ \midrule
        VAR & \textbf{1.131} & 968048 & 10.30 & 22.11 & 17.87 \\
        GB Forecaster & 1.705 & \textbf{41.66} & \textbf{3.712} & \textbf{14.05} & \textbf{12.18}   \\
        RF Forecaster & 3.256 & 44.20 & 4.633 & 23.25 & 19.82\\
        \bottomrule
    \end{tabular}
    \caption{Performance of multivariate forecasting models. All models were evaluated without retraining. Best results are in \textbf{bold}. We report RMSE instead of sMAPE for the Seattle Trail and Solar Plant datasets because a large portion of the data values (13\% and 55\% respectively) are equal to 0, making them unsuitable for sMAPE. 1 TS means that there is only one time series in the dataset.}
    \label{tab:multi_forecast_result}
\end{table}

\subsection{Univariate Anomaly Detection}
\label{sec:experiments_uni_anom}

\subsubsection{Datasets and Evaluation}
\begin{table}
    \small
    \centering
    \begin{tabular}{r|c|c|c|c}
        \toprule
        & Int\_UA & NAB & AIOps & UCR \\ \midrule
        \# Time Series & 26 & 58 & 29 & 250 \\
        Real data? & \cmark & 47/58 real & \cmark & Mixed \\
        Availability & Internal & Public & Public & Public \\ 
        Data Source & Cloud KPIs & Various & Cloud KPIs & Various \\
        \midrule
        Train split & First 25\% & First 15\% & First 50\% & First 30\% \\
        Supervised? & No & No & Train labels & Test labels \\
        Threshold & 4.0 & 3.5 & -- & -- \\
        \midrule
        Reference & -- & {\footnotesize \cite{nab}} & {\footnotesize \cite{aiops_challenge}} & {\footnotesize \cite{UCRArchive2018}} \\
        \bottomrule
    \end{tabular}
    \caption{Summary of univariate anomaly detection datasets.}
    \label{tab:uni_anom_data}
\end{table}

We report results on four public and internal datasets (Table \ref{tab:uni_anom_data}). For the internal dataset and the Numenta Anomaly Benchmark (NAB, \cite{nab}), we choose a single (calibrated) detection threshold for all time series and all algorithms. For the AIOps challenge \citep{aiops_challenge}, we use the labeled anomalies in the training split of each time series to choose the detection threshold that optimizes F1 on the training split; for the UC Riverside Time Series Anomaly Archive \citep{UCRArchive2018}, we choose the detection threshold that optimizes F1 on the test split. This is because the dataset is incredibly diverse (so a single threshold doesn't apply to all time series), and there are no anomalies present in the training split. Table \ref{tab:uni_anom_data} summarizes these datasets and evaluation choices.

We use the evaluation pipeline described in \S\ref{sec:evaluation} to evaluate each model. After training an initial model on the training split of a time series, we re-train the model unsupervised either daily or hourly on the full data until that point (without adjusting the calibrator or threshold). We then {\em incrementally} obtain predicted anomaly scores for the full time series, in a way that simulates a live deployment scenario. We also consider batch prediction, where the initial trained model predicts anomaly scores for the entire test split in a single step, without any re-training. Note that the UCR dataset does not contain timestamps; we treat it as if it is sampled once per minute, but this is an imperfect assumption. We consider only batch prediction and ``daily'' retraining for this dataset for efficiency reasons.

\subsubsection{Models}
We evaluate two classes of models: forecast-based anomaly detectors and statistical methods. For forecast-based methods, we use ARIMA, AutoETS, and AutoProphet (as described in \S\ref{sec:experiments_uni_forecast}). For statistical methods, we use Isolation Forest \citep{isolation_forest}, Random Cut Forest \citep{random_cut_forest}, and Spectral Residual \citep{spectral_residual}, as well as two simple baselines WindStats and ZMS (\S\ref{sec:anom_models}). We also consider an ensemble of AutoETS, RRCF, and ZMS (using the algorithm described in \S\ref{sec:anom_ensemble}).

\subsubsection{Results}
\begin{table}
    \small
    \centering
    \begin{tabular}{r|cccc|c}
    \toprule
    {} &       Int\_UA &           NAB &         AIOps &           UCR &  $\Delta$ F1 (vs. best) \\
    \midrule
    ARIMA                &  $\bm{0.531}$ &       $0.395$ &       $0.227$ &       $0.313$ &       $0.148 \pm 0.099$ \\
    AutoETS              &       $0.296$ &       $0.350$ &       $0.097$ &       $0.334$ &       $0.245 \pm 0.042$ \\
    AutoProphet          &       $0.343$ &       $0.323$ &       $0.310$ &       $0.418$ &       $0.166 \pm 0.062$ \\
    \midrule
    Isolation Forest     &       $0.436$ &       $0.244$ &       $0.347$ &       $0.461$ &       $0.142 \pm 0.111$ \\
    Random Cut Forest    &       $0.248$ &       $0.337$ &       $0.314$ &  $\bm{0.568}$ &       $0.148 \pm 0.132$ \\
    Spectral Residual    &       $0.340$ &       $0.153$ &       $0.338$ &       $0.469$ &       $0.189 \pm 0.150$ \\
    WindStats (baseline) &       $0.225$ &       $0.247$ &       $0.324$ &       $0.306$ &       $0.239 \pm 0.114$ \\
    ZMS (baseline)       &       $0.486$ &       $0.290$ &       $0.340$ &       $0.427$ &       $0.129 \pm 0.094$ \\
    \midrule
    Ensemble (ours)      &       $0.500$ &  $\bm{0.548}$ &  $\bm{0.396}$ &       $0.476$ &  $\bm{0.034 \pm 0.044}$ \\
    \bottomrule
    \end{tabular}
    \caption{F1 scores achieved by univariate anomaly detection models. All models were evaluated using batch prediction, daily re-training, and hourly re-training; we report the best F1 achieved by any of the re-training schedules. We also report the average gap (over datasets) in F1 between each model and the best model. Best results are in \textbf{bold}.}
    \label{tab:uni_anom_full_results}
\end{table}

\begin{table}
    \small
    \centering
    \begin{tabular}{r|cc}
    \toprule
    {} &                   Daily &                  Hourly \\
    \midrule
    ARIMA                &   $0.162$ ($p = 0.001$) &   $0.227$ ($p < 0.001$) \\
    AutoETS              &   $0.088$ ($p = 0.017$) &   $0.058$ ($p = 0.325$) \\
    AutoProphet          &   $0.028$ ($p = 0.013$) &                      -- \\
    \midrule
    Isolation Forest     &   $0.006$ ($p = 0.779$) &   $0.019$ ($p = 0.567$) \\
    Random Cut Forest    &   $0.024$ ($p = 0.396$) &   $0.033$ ($p = 0.337$) \\
    Spectral Residual    &  $-0.090$ ($p = 0.075$) &  $-0.109$ ($p = 0.004$) \\
    WindStats (baseline) &   $0.044$ ($p = 0.098$) &   $0.017$ ($p = 0.145$) \\
    ZMS (baseline)       &   $0.005$ ($p = 0.929$) &   $0.011$ ($p = 0.872$) \\
    \midrule
    Ensemble (ours)      &   $0.147$ ($p = 0.023$) &   $0.183$ ($p = 0.009$) \\
    \bottomrule
    \end{tabular}
    \caption{Average change in F1 (relative to no re-training) achieved by re-training models both daily and hourly. $p$-value is from a 2-sided paired sample $t$-test. Re-training makes a significant improvement for the forecasting models (first block) and the ensemble. However, the statistical models (second block) benefit only marginally (if at all).}
    \label{tab:uni_anom_retrain}
\end{table}
Table \ref{tab:uni_anom_full_results} reports the revised point-adjusted F1 score achieved by each model on each dataset. We report the best F1 score achieved by any of the 3 re-training schedules (for efficiency reasons, we only consider batch prediction and daily re-training for AutoProphet). We find that our proposed ensemble of AutoETS, RRCF, and ZMS achieves the best performance on two of the four datasets, and the second-best on the others; overall, it has the smallest average gap in F1 score relative to the best model for each dataset (along with the smallest variance in this gap). For this reason, we believe that it is a good ``default'' model for new users or early exploration.

Table \ref{tab:uni_anom_retrain} examines the impact of the re-training schedule for each model, averaged across all datasets. We find that daily and hourly re-training of forecasting-based models and our proposed ensemble can greatly improve their anomaly detection performance. However, in practice, the ideal re-training frequency may require some experimentation to choose. Interestingly, the impact of frequent re-training is unclear for most statistical models.

\subsection{Multivariate Anomaly Detection}
\label{sec:experiments_multi_anom}
\subsubsection{Datasets and Evaluation}
The evaluation setting for multivariate anomaly detection is nearly identical to that for univariate anomaly detection. Table \ref{tab:multi_anom_data} describes the multivariate time series anomaly detection datasets in this epxeriment. Note that we treat anomaly detection on all these datasets as a fully unsupervised learning task. The main difference from the univariate setting is that we consider only batch predictions and weekly retraining (rather than batch predictions, daily re-training, and hourly retraining) for efficiency reasons.

\begin{table}
    \small
    \centering
    \begin{tabular}{r|c|c|c|c}
        \toprule
        & SMD & SMAP & MSL & Int\_MA \\ \midrule
        \# Time Series & 28 & 1 & 1 & 20 \\
        \# Variables & 38 & 25 & 55 & 5 \\
        Real data? & \cmark & \cmark & \cmark & \cmark \\
        Availability & Public & Public & Public & Internal \\ 
        Data Source & Cloud KPIs & Satellite Data & Mars Rover & Cloud KPIs \\ \midrule
        Train split & First 50\% & First 25\% & First 45\% & First 50\% \\
        Supervised? & No & No & No & No \\
        Threshold & 3.0 & 3.5 & 3.0 & 3.5 \\
        \midrule
        Reference & {\footnotesize \cite{smd}} & {\footnotesize \cite{hundman2018detecting}} & {\footnotesize \cite{hundman2018detecting}} & -- \\
        \bottomrule
    \end{tabular}
    \caption{Summary of multivariate anomaly detection datasets.}
    \label{tab:multi_anom_data}
\end{table}

\subsubsection{Models}
We consider two classes of models: statistical methods and deep learning models. For statistical methods, we evaluate Isolation Forest \citep{isolation_forest} and Random Cut Forest \citep{random_cut_forest} as in \S\ref{sec:experiments_uni_anom}. For deep learning approaches, we evaluate an autoencoder \citep{autoencoder}, a Deep Autoencoding Gaussian Mixture Model (DAGMM, \cite{dagmm}), a LSTM encoder-decoder \citep{lstmed}, and a variational autoencoder \citep{vae}. We also consider an ensemble of a Random Cut Forest and a Variational Autoencoder (using the algorithm described in \S\ref{sec:anom_ensemble}).

\subsubsection{Results}
\begin{table}
    \small
    \centering
    \begin{tabular}{r|cccc|c}
    \toprule
    {} &           SMD &          SMAP &           MSL &       Int\_MA &  $\Delta$ F1 (vs. best) \\
    \midrule
    Isolation Forest        &       $0.388$ &       $0.211$ &       $0.226$ &       $0.252$ &       $0.129 \pm 0.025$ \\
    Random Cut Forest       &  $\bm{0.500}$ &       $0.108$ &       $0.312$ &       $0.313$ &       $0.091 \pm 0.110$ \\
    \midrule
    Autoencoder             &       $0.264$ &  $\bm{0.357}$ &       $0.355$ &  $\bm{0.357}$ &       $0.065 \pm 0.114$ \\
    DAGMM                   &       $0.191$ &       $0.037$ &       $0.135$ &       $0.182$ &       $0.262 \pm 0.067$ \\
    LSTM Encoder-Decoder    &       $0.344$ &  $\bm{0.357}$ &  $\bm{0.381}$ &       $0.301$ &       $0.053 \pm 0.074$ \\
    Variational Autoencoder &       $0.318$ &       $0.187$ &  $\bm{0.381}$ &       $0.322$ &       $0.097 \pm 0.093$ \\
    \midrule
    Ensemble (ours)         &       $0.403$ &       $0.291$ &       $0.357$ &       $0.341$ &  $\bm{0.051 \pm 0.038}$ \\
    \bottomrule
    \end{tabular}
    \caption{F1 scores achieved by multivariate anomaly detection models. All models were evaluated using batch prediction and weekly re-training; we report the best F1 achieved by any of the re-training schedules. We also report the average gap (over datasets) in F1 between each model and the best model. Best results are in \textbf{bold}.}
    \label{tab:multi_anom_full_results}
\end{table}

\begin{table}
    \small
    \centering
    \begin{tabular}{r|c}
    \toprule
    {} &                  Weekly \\
    \midrule
    Isolation Forest        &   $0.006$ ($p = 0.796$) \\
    Random Cut Forest       &   $0.036$ ($p = 0.130$) \\
    \midrule
    Autoencoder             &  $-0.075$ ($p = 0.352$) \\
    DAGMM                   &   $0.057$ ($p = 0.032$) \\
    LSTM Encoder-Decoder    &   $0.001$ ($p = 0.970$) \\
    Variational Autoencoder &   $0.020$ ($p = 0.286$) \\
    \midrule
    Ensemble (ours)         &  $-0.032$ ($p = 0.499$) \\
    \bottomrule
    \end{tabular}
    \caption{Average change in F1 (relative to no re-training) achieved by re-training models weekly. $p$-value is from a 2-sided paired sample $t$-test. The changes are not statistically significant at $p = 0.05$ for any models besides DAGMM.}
    \label{tab:multi_anom_retrain}
\end{table}

Table \ref{tab:multi_anom_full_results} reports the revised point-adjusted F1 score achieved by each model. While there is no clear winner across all datasets, our proposed ensemble consistently achieves a small gap in F1 score relative to the best model for each dataset. The LSTM encoder-decoder achieves a similar average gap, but it has a larger variance. Therefore, we believe the ensemble to be a good ``default'' model for new users or early exploration, as it robustly obtains reasonable performance across multiple datasets. Finally, like the univariate case, Table \ref{tab:multi_anom_retrain} shows that the impact of re-training is ambiguous for all the multivariate statistical models considered (first block), as well as the deep learning models (second block).

\section{Conclusion and Future Work}
We introduce Merlion, an open source machine learning library for time series, which is designed to address many of the pain points in today's industry workflows for time series anomaly detection and forecasting. It provides unified, easily extensible interfaces and implementations for a wide range of models and datasets, an autoML module that consistently improves the performance of multiple forecasting models, post-processing rules for anomaly detectors that improve interpretability and reduce the false positive rate, and transparent support for ensembles that robustly achieve good performance on multiple benchmark datasets. These features are tied together in a flexible pipeline that quantitatively evaluates the performance of a model and a visualization module for more qualitative analysis.

We continue to actively develop and improve Merlion. Planned future work includes adding support for more models including the latest deep learning models and online learning algorithms, developing a streaming platform to facilitate model deployment in a real production environment, and implementing advanced features for multivariate time series analysis. We welcome and encourage any contributions from the open source community.

\section*{Acknowledgements}
We would like to thank a number of leaders and colleagues from Salesforce who have provided strong support, advice, and contributions to this open-source project. 

\vskip 0.2in
\bibliography{references}

\end{document}